%% file: ref.tex
\def\year{2021}\relax
\documentclass[letterpaper]{article} 
\usepackage{aaai21}  
\usepackage{times}  
\usepackage{helvet} 
\usepackage{courier}  
\usepackage[hyphens]{url}  
\usepackage{graphicx} 
\usepackage{natbib}  
\usepackage{caption} 
\usepackage{color}
\usepackage{xspace}
\usepackage{multirow}
\usepackage{subcaption}
\usepackage{array}
\usepackage{booktabs}
\usepackage{amsmath}
\usepackage[dvipsnames]{xcolor}

\newif\ifsubmit

\submittrue

\newcommand{\ie}[0]{\emph{i.e.,}\xspace}

\newcommand{\eg}[0]{\emph{e.g.,}\xspace}

\newcommand{\sys}[0]{Inter-Group Robustness Prioritization\xspace}
\newcommand{\sysshort}[0]{IGRP\xspace}

\title{Adaptive Verifiable Training Using Pairwise Class Similarity}

\author {
    Shiqi Wang\textsuperscript{\rm 1},
    Kevin Eykholt\textsuperscript{\rm 2},
    Taesung Lee\textsuperscript{\rm 2},
    Jiyong Jang\textsuperscript{\rm 2},
    Ian Molloy\textsuperscript{\rm 2} \\
}
\affiliations {
    \textsuperscript{\rm 1} Columbia University \\
    \textsuperscript{\rm 2} IBM Research \\
    tcwangshiqi@cs.columbia.edu; \{kheykholt, Taesung.Lee\}@ibm.com; \{jjang, molloyim\}@us.ibm.com
}

\begin{document}

\maketitle

 \begin{abstract}
 \input{section/abstract}
 \end{abstract}
 

 \input{section/introduction}
 

 \input{section/background}
 

 \input{section/design}
 

 \input{section/results}


 \input{section/conclusion}

 \input{section/ethical_impact}

\begin{quote}
\begin{small}
\bibliography{ref}
\end{small}
\end{quote}

\input{section/appendix}

\end{document}

%% file: section/abstract.tex
Verifiable training has shown success in creating neural networks that are provably robust to a given amount of noise. However, despite only enforcing a single robustness criterion, its performance scales poorly with dataset complexity. On CIFAR10, a non-robust LeNet model has a 21.63\% error rate, while a model created using verifiable training and a $L_{\infty}$ robustness criterion of 8/255, has an error rate of 57.10\%. Upon examination, we find that when labeling visually similar classes, the model's error rate is as high as 61.65\%. Thus, we attribute the loss in performance to inter-class similarity. Classes that are similar (\ie close in the feature space) increase the difficulty of learning a robust model. While it may be desirable to train a model to be robust for a large robustness region, pairwise class similarities limit the potential gains. Furthermore, consideration must be made regarding the relative cost of mistaking one class for another. In security or safety critical tasks, similar classes are likely to belong to the same group, and thus are equally sensitive.

In this work, we propose a new approach that utilizes inter-class similarity to improve the performance of verifiable training and create robust models with respect to \emph{multiple adversarial criteria}. First, we cluster similar classes using agglomerate clustering and assign robustness criteria based on the degree of similarity between clusters.  Next, we propose two methods to apply our approach: (1) the \sys method, which uses a custom loss term to create a single model with multiple robustness guarantees and (2) the neural decision tree method, which trains multiple sub-classifiers with different robustness guarantees and combines them in a decision tree architecture. Our experiments on Fashion-MNIST and CIFAR10 demonstrate that by prioritizing the robustness between the most dissimilar groups, we improve clean performance by up to 9.63\% and 30.89\% respectively. Furthermore, on CIFAR100, our approach reduces the clean error rate by 26.32\%.

%% file: section/introduction.tex
\label{sec:introduction}
\section{Introduction}

Dramatic improvements in the accuracy of neural networks on various tasks has been made, but their robustness is often not prioritized. However, with poor robustness, the security and reliability of models is in question when exposed to adversarial examples. Despite appearing indistinguishable from a normal input, adversarial examples consistently induce predictable errors in machine learning models. While many defensive techniques have been developed, most fall short as they obfuscate the discovery process rather that truly reducing the number of adversarial examples a model is vulnerable to. One effective defense against adversarial examples is verifiable training as it creates models with provable robustness guarantees. With respect to a robustness criteria, which identifies a region around an input where the model's prediction must remain stable, verifiable training maximizes the potential number of input samples a model is certified to be robust for within that region.

Although verifiable robust training creates models with provable robustness, it often comes at the cost of lower performance on clean data. For example, on CIFAR10, a LeNet model trained using CROWN-IBP, a state-of-the-art verifiable training method, with respect to a $L_{\infty}$ robustness region $\epsilon = \frac{8}{255}$,
has significantly lower clean performance compared to a model created through normal training (\ie $57.10\%$ error rate vs. $21.53\%$ error rate).
With such poor baseline performance, certified performance on adversarial samples is limited, having only a $69.92\%$ error rate.
This means that in presence of an adversary, only about $30\%$ of the inputs are guaranteed to be correctly classified.
\newline

\begin{figure}[h]
\includegraphics[width=\columnwidth]{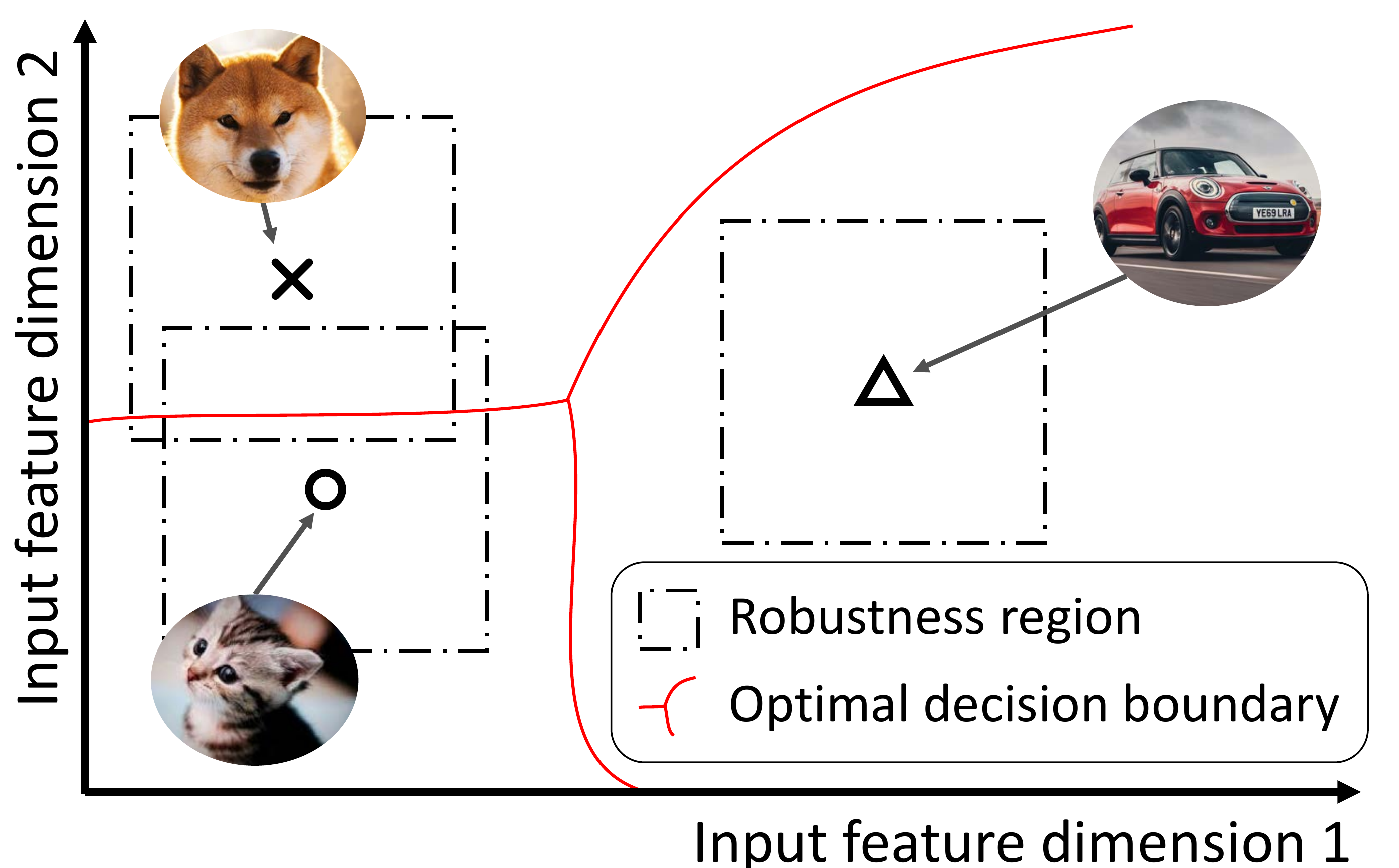} 
\caption{During verification, inputs belonging to similar classes may have overlapping robustness regions  when the robustness region is too large.}

\label{fig:similarity}
\end{figure}

The poor performance of existing verifiable training methods is due to using only a single robustness criteria.
During verification, the robustness region around an input is estimated and used to determine if the decision is stable within the region. Verifiable training attempts to shape the decision boundary so as to maximize the number of inputs the model's decision is stable for within the robustness region.
However, as shown in Figure~\ref{fig:similarity}, inputs belonging to similar classes may have overlapping estimations of their robustness regions, thus resulting in high confusion between these classes.
For example, a robust LeNet model trained on CIFAR10 mislabels a dog as a cat 33.53\% of the time, whereas it mislabels a dog as a car only 7.48\% of the time.
These inherent inter-class similarities in the data limit the natural performance of verifiable training if only a single robustness criteria is used.
Additionally, the inter-class similarity can also represent the relative sensitivity cost of a misclassification.
In safety or security critical tasks, the cost of misclassifying similar classes is likely lower than the cost of misclassifying dissimilar ones. In autonomous driving, misidentifying a Speed Limit 40 sign as a Speed Limit 30 causes the car to change its speed only.
However, misidentifying a Speed Limit sign as a Stop sign causes the car to come to a sudden halt. Such cost-sensitive situations naturally encourage using different robustness criteria during training based on inter-class similarity.

In this paper, we propose \textit{adaptive verifiable training}, enabling us to create machine learning models \textit{robust with respect to multiple robustness criteria}. Because the current state-of-the-art is limited to using a single robustness criteria, all data classes are treated equally. As such, for strict robustness criteria (\ie high $\epsilon$), the overall performance of a model degrades due to conflicting estimations of robustness regions between similar classes. To address this issue, adaptive verifiable training exploits the inherent inter-class similarities within the data and enforces multiple robustness criteria based on this information. Between similar classes, our approach enforces looser robustness criteria (\ie smaller $\epsilon$) so as to minimize any possible overlap when estimating the robustness region during verification. Between dissimilar classes, on the other hand, our approach enforces stricter robustness regions (\ie larger $\epsilon$). Our contributions are:

\begin{itemize}
    \item A novel approach that allows the state-of-the-art verifiable training techniques to create models robust to more than just a single robustness region. We exploit pairwise class similarities, thus improving the performance of robust models by relaxing the robustness constraints for similar classes and increasing the robustness constraints for dissimilar classes.
    \item A method to automatically identify and cluster similar and dissimilar classes based on prior work.
    \item Two methods to create classifiers for multiple robustness criteria: (1) \sys which uses a single model architecture and encodes multiple robustness requirements using a customized loss function, and; (2) neural decision trees which creates a robust model using a decision tree architecture where each node in the tree is an individually robust classifier that identifies the class group a given input belongs to.
    \item We perform an empirical evaluation comparing our approach to CROWN-IBP, a state-of-the-art verifiable training technique. Adaptive verifiable training, which prioritizes the model's robustness against misclassification between the most dissimilar classes, results in lowering the clean error rate for a large value of $\epsilon$ by 9.63\% and  30.89\% on Fashion-MNIST and CIFAR10, respectively. On CIFAR100, we achieve up to 26.32\% reduction in error rate. In all cases, the robustness of the model with respect to the dissimilar classes is preserved.
\end{itemize}

%% file: section/background.tex
\label{sec:background}
\section{Background and Related Work}

As neural networks are used in critical applications, such as autonomous driving and medical diagnosis, it is important to ensure that the trained models are robust and trustworthy.
Recently, researchers found that most state-of-the-art models are not robust in the presence of an adversary. \citet{szegedy2014intriguing} classified the existence of  \textbf{adversarial examples} in neural networks as input samples with carefully crafted \textit{imperceptible} noise
that induce predictable and transferable misclassification errors. Their work inspired the design of many new attack algorithms that adversarial examples under different threat models~\cite{papernot2016limitations,goodfellow2014explaining,kurakin2016adversarial,zoo,Bhagoji2018PracticalBA,Cheng2018QueryEfficientHB,papernot2016transferability,liu2016delving}. In response, many defensive measures have been proposed, but without a provable guarantees of performance.
\citet{athalye2018obfuscated} showed that many of these defenses rely on shattering the model's loss gradient and could be successfully attacked by making small modifications to existing attacks.

A promising direction is verifiable training. Unlike the defenses analyzed by \citet{athalye2018obfuscated} which only provide empirical proof of a successful defense, verifiable training provides theoretical guarantees of a model's robustness.
For each input sample, verifiable training computes the lower bound logit value of the true class and the upper bound logit values of the other classes with respect to a given robustness criteria ($\epsilon$). The bounds can be computed using verification algorithms like Interval Bound Propagation (IBP)~\cite{gowal2018effectiveness}, or other more expensive linear relaxation methods~\cite{zhang2018efficient,wong2018provable,weng2018towards,wang2018efficient,singh2018fast}. We refer the readers to the works by \citet{salman2019convex} and \citet{liu2019algorithms} for a comprehensive survey on verification algorithms. The model is updated such that the computed lower bound of the true class is always greater than the upper bound of other classes~\cite{wong2018provable}. However, the current state-of-the-art only provides theoretical guarantees for a single robustness region. 

Our proposed approach extends verifiable training methods to provide provable guarantees for multiple robustness regions. We leverage pairwise class similarity to build adaptive robustness regions, thus maximizing the performance of robust models. Utilizing the pairwise class relationships in data has been done prior works. \citet{zhang2018cost} define the cost of misclassifications based on class relationships. They define an $N \times N$ cost matrix for $N$ input classes, and minimize a weighted loss function, which is the product of the cost matrix and a standard loss function. Their approach, which focuses on the robustness of certain classes of interest, is orthogonal to ours, as we aim to apply different robust regions using class similarity and relax the unsatisfiable conditions. Another related work is DL2~\cite{fischer2019dl2}, which enforces logical constraints during the training process. However, their approach relies on empirical adversarial training to enforce these constraints, thus they cannot establish provable robustness guarantees.

In regards to network ensembles, \citet{jonas2020certifying} and \citet{zhang2019enhancing} both consider creating an ensemble of classifiers to improve adversarial robustness. Both works focus on training diverse component classifiers~\citep{jonas2020certifying},
or finding the optimal ensemble weights and training complementary components~\citep{zhang2019enhancing}. Our approach differs in that we use ensembling to improve the clean performance of a robust model. Furthermore, we construct our ensemble differently. Their designs follows traditional ensembling methods, in which each model performs the same task as the other models, whereas our design use a decision tree structure. Our approach can be combined with theirs when creating robust ensembles.

Finally, our approach is designed independent of the verification method. Although we use CROWN-IBP~\cite{zhang2019towards} enhanced with loss fusion~\cite{xu2020automatic}, where verifiable robustness is measured using IBP, other verifiable training methods can be combined with our framework and our approach would remain valid ~\cite{wang2018mixtrain,wong2018scaling,raghunathan2018certified,mirman2018differentiable,gowal2018effectiveness, balunovic2020Adversarial}. Randomized smoothing combined with noisy data augmentation \cite{cohen2019certified, salman2019provably} is another popular technique to create provably robustness guarantees by transforming a pre-trained classifier into a new smoothed classifier. Given an input, the smoothed classifier generates multiple noisy corruptions of the input and outputs the most likely prediction. This approach can also be used with our proposed framework as randomized smoothing augments pre-trained classifiers. We leave it for future work.

%% file: section/design.tex
\section{Adaptive Verifiable Training}
\label{sec:design}
In this section, we provide an overview of adaptive verifiable  training, our  methodology for  creating  machine learning models with multiple robustness certificates. Existing verifiable training creates a model that is only robust with respect to a  single robustness criteria  based  on  the  assumption  that  all  errors  are  equal. However, we argue that certain errors made by the model, whether due to natural error or adversarial manipulation, are easier to make due to the inherent similarities between classes. Classes that are highly similar (e.g., dogs and cats) limit the model performance when the robustness criteria is overly  strict  due  to  overlapping  robustness  regions  during verification. Our approach addresses this problem by creating models with relaxed robustness criteria between similar classes, while maintaining strict robustness criteria between  dissimilar  classes.  First,  we identify  the  inter-class relationships and define robustness criteria to enforce with respect to these relationships. Once defined, we enforce the robustness constraints either using \sys or Neural Decision Trees.

\subsection{Class Similarity Identification}
The first step of our approach is to identify similar class pairs and infer which relationships should have relaxed robustness constraints. Absent predefined pairwise class relationships, we propose using agglomerative clustering to define the similarity between classes. Given the weights of the penultimate  layer  of  a  pre-trained  classifier, agglomerative clustering pairs classes together based on a similarity metric (\eg $L_{2}$ distance). After creating the initial clusters, the process iteratively combines smaller clusters into larger clusters using the same similarity metric, until only a single cluster remains.  A similar approach was used by \citet{wan2020nbdt} where they replaced the final layer of a neural network with a decision tree to provide explainability around the network’s predictions. Once the classes have been clustered, we need to define the robustness criteria to certify a model against for each group. The robustness criteria, $\epsilon$, can be determined by the user, and in general the robustness criteria can be increased increase as the class similarity decreases.

\subsection{\sys (\sysshort)}
The \sys (\sysshort) method follows traditional verifiable training and only creates a single robust model. Unlike prior work, a model created using \sysshort can enforce multiple robustness criteria based on the class grouping during similarity identification. Here, we describe traditional verifiable training and discuss the improvements of our method.

\noindent{\bf Verification specification and verifiable robustness.}
In neural network verification, the verification specification for an  input  sample $x_k$ is defined by a  specification  matrix $\boldsymbol{C} \in \rm I\!R^{n_L \times n_L}$, where $n_L$ is the number of classes. Given the true label $y$, we define the specification matrix as:

\begin{equation}
    C_{i,j} = \begin{cases}
                1 &\text{if $j=y, i\neq y$}\\
                -1 &\text{if $i=j, i\neq y$}\\
                0 &\text{otherwise}
            \end{cases}
\end{equation}
Thus, for each row vector $\boldsymbol{c_i} \in \rm I\!R^{n_L}$ in the specification matrix, the index of the true label is 1, the index of the current label is -1, and all other indices are 0. For the row vector $c_y$ corresponding to the true label, all indices are 0.

We use the above definition to define the margin vector $\boldsymbol{m}(\boldsymbol{x}):=\boldsymbol{C}\boldsymbol{f}(\boldsymbol{x}) \in \rm I\!R^{n_L}$ where each element $m_i$ in the margin vector denotes the margin between class $y$ and the other classes $i$ (\eg $f_{y}(x) - f_{i}(x))$. Next, given the robustness region $S(x_{k}, \epsilon) = \{x:||x_k - x||_p \leq \epsilon\}$, we define the lower bound of $\boldsymbol{C}\boldsymbol{f}(\boldsymbol{x})$ for all $x \in S(x, \epsilon)$ as $\boldsymbol{\underline{m}}(\boldsymbol{x_{k}},\epsilon)$. The values in $\boldsymbol{\underline{m}}(\boldsymbol{x_{k}},\epsilon)$ represent the worst-case margin values for the input. When all elements  in $\boldsymbol{\underline{m}}(\boldsymbol{x_{k}}, \epsilon) > 0$, $x_k$ is \emph{verifiably  robust} for  any perturbation  in $S(x_{k}, \epsilon)$.  Verification  methods  like  IBP  can be used to obtain $\boldsymbol{\underline{m}}(\boldsymbol{x_{k}}, \epsilon)$.

\noindent{\bf Verifiable training.} The min-max robust optimization widely used in adversarial training is defined as :
\begin{equation}
   \min_{\theta} E_{(\boldsymbol{x}, y)\in \mathcal{D}} \Big[ \max_{x \in S(x_{k}, \epsilon)}L(\boldsymbol{f}(\boldsymbol{x});y;\theta)
   \Big]
\label{eq:adv_training}
\end{equation}

Due to the non-linearity of neural networks, the inner maximization problem is a challenging problem to solve. Rather than solve the inner maximization problem, \citet{wong2018provable} showed that the worst-case margin vector can serve as sound upper bound, \ie:
\begin{equation}
    \max_{x \in S(x_{k}, \epsilon)} L(\boldsymbol{f}(\boldsymbol{x});y;\theta)\leq L(-\boldsymbol{\underline{m}}(\boldsymbol{x}_{k}, \epsilon);y;\theta)
\label{eq:upperbound}
\end{equation}
Traditional verifiable training uses Equation \ref{eq:upperbound} and trains model to minimize this upper bound for the inner maximization for each training input. This maximizes the performance with respect to $\epsilon$, resulting in a verifiably robust model.

\noindent{\bf \sysshort.} In order to support multiple robustness criteria, \sysshort defines two loss terms: the outer group loss and the inner group loss. Given a set of class groups ${G_1, G_2...G_k}$, an input $x_{k}$, and the true label $y$, the \textbf{outer group loss}, $L_{outer}$, is defined as the loss between the group the true label belongs to, $G^y$,  and the other groups. When computing the worst-case margin values, classes within the same group as the true label $y$ are not considered. We enforce this by zeroing them out. Formally, the verification specification matrix, $C^O_{i,j}$, for the outer loss is defined as:

\begin{equation}
    C^O_{i,j} = \begin{cases}
                1 &\text{if $j=y, i\neq y$, $G^i \neq G^y$}\\
                -1 &\text{if $i=j, i\neq y$, $G^i \neq G^y$}\\
                0 &\text{otherwise}
            \end{cases}
    \label{eq:louter}
\end{equation}

\noindent The margin vector for the outer robustness criteria is defined as $\boldsymbol{m^O}(\boldsymbol{x})=\boldsymbol{C^Of}(\boldsymbol{x})$ and the outer loss is defined as $L_{outer}=L(\boldsymbol{-\underline{m}^O}(\boldsymbol{x}_{k},\epsilon^O);y;\theta)$.

Similarly, the \textbf{inner group loss}, $L_{inner}$, is defined as the loss between labels belonging to the same group as the true label. When computing the worst-case margin values, classes that are in a different group as the true label $y$ are not considered. We enforce this by zeroing them out. Formally, the verification specification matrix, $C^I_{i,j}$, for the inner loss is defined as:

\begin{equation}
    C^I_{i,j} = \begin{cases}
                1 &\text{if $j=y, i\neq y$, $G^i = G^y$}\\
                -1 &\text{if $i=j, i\neq y$, $G^i = G^y$}\\
                0 &\text{otherwise}
            \end{cases}
    \label{eq:linner}
\end{equation}

\noindent The margin vector for the inner robustness criteria is defined as $\boldsymbol{m^I}(\boldsymbol{x})=\boldsymbol{C^If}(\boldsymbol{x})$ and the inner loss is defined as $L_{inner} = L(\boldsymbol{-\underline{m}^I}(\boldsymbol{x}_{k},\epsilon^I);y;\theta)$.

Given the definitions of $L_{outer}$ and $L_{inner}$, we define the \sysshort training objective as:

\begin{equation}
    L_{\sysshort} = L_{outer}+L_{inner}
    \label{eq:new_loss}
\end{equation}

\noindent By using verifiable training to minimize Equation \ref{eq:new_loss}, adaptive verifiable training creates a single robust model with respect to multiple robustness criteria. Dissimilar classes are clustered into different groups, so we use outer loss term to enforce a strict robustness criteria between those groups. Similar classes are clustered into the same group, so we use the inner loss term to enforce a loose robustness criteria between those groups. Furthermore, if multiple outer or inner group relationships exist, we can simply add a new inner or outer loss term to $L_{IGRP}$. Note that the computational cost of \sysshort is theoretically the same as traditional verifiable training as we only need to estimate the worst-case margin value for each class once during verification even though multiple robustness distances may be considered.

\subsection{Neural Decision Tree (NDT)}
A Neural Decision Tree (NDT) is a decision tree where each tree node is a neural network classifier. By training each node in the tree using a different value of $\epsilon$, we can enforce multiple robustness constraints. Once classes have been clustered together, we train each node to identify the group or subgroup an input sample belongs to. As an input is passed through the tree, the model's output become more fine-grained, predicting groups with fewer labels. The final prediction of the NDT is made when only a single class label is predicted. For example, after using agglomerative clustering on CIFAR10 with a binary split, the root node determines if an input belongs to [bird, cat, dog, deer, frog, horse] or [airplane, car, ship, truck]. Let's assume the right child is always predicted. The next node classifies predicts if the input belongs to [airplane, ship] or [car, truck]. Finally, the final node predicts if the input is either a car or a truck.

\begin{figure}[h]
\includegraphics[width=\columnwidth]{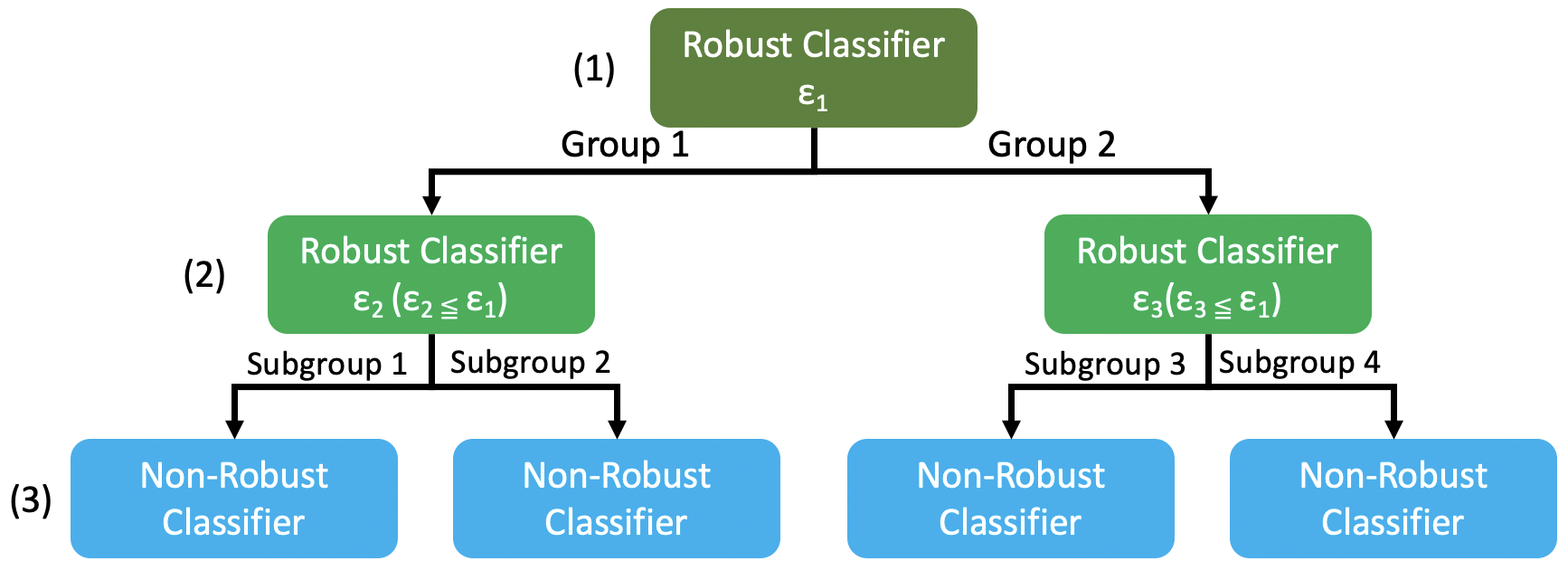}
\caption{A basic NDT architecture. Given an input: (1) Each node determines which of two groups of class labels the in-put belongs to; (2) Once identified, input is passed to the next respective model in the tree; (3) Finally, at the leaf nodes, if the predicted subgroup only contains a single label, a final classification is output. The predicted class is determined by the path of the input through the NDT.}
\label{fig:ndt}
\end{figure}

As each node in the tree is distinct, we can easily support multiple robustness criteria depending on the similarity of the groups at a particular node. The only requirement is that the parent must be at least as robust as its children. Generally, the closer a node is to the root of the tree, the stricter the robustness criteria can be as the similarity between groups decreases. We note that although Figure \ref{fig:ndt} shows a tree with a mix of binary robust and non-robust classifiers, our approach is not limited to this construction. In Section \ref{sec:results}, we present results using several constructions architectures, including some where we combine binary and non-binary models to achieve high performance on CIFAR100.

%% file: section/results.tex
\section{Results}
\label{sec:results}

\subsection{Evaluation Setup}

\noindent{\bf Datasets.} We evaluated our approach using the Fashion-MNIST (F-MNIST), CIFAR10, and CIFAR100 datasets. For evaluation on F-MNIST and CIFAR10, we used two different $L_\infty$ norms,  $\epsilon=[0.1, 0.3]$ and $\epsilon=[\frac{2}{255}, \frac{8}{255}]$, respectively. For evaluation on CIFAR100, due to the poor performance of the baseline model, we only used $\epsilon=\frac{2}{255}$.

\noindent{\bf Models.} We experimented with several different model architectures, but found that the performance difference between our baseline comparison model and our approach did not significantly change across model architectures. Our F-MNIST and CIFAR10 results were generated using a 4-layer LeNet model, which was used in prior work~\cite{zhang2019towards,gowal2018effectiveness}. In the full paper, we include the results for a second larger LeNet model, denoted DM-Large in prior work. For CIFAR100, we switched to the ResNet model\footnote{In order to support this architecture, we use the AutoLIRPA library~\cite{xu2020provable}}, due to better baseline performance. Our Neural Decision Trees use the same model architecture for each node as the baseline model we compare against.

\noindent{\bf Training and Evaluation.} We used the state-of-the-art verifiable training method CROWN-IBP~\cite{zhang2019towards} enhanced with loss fusion~\cite{xu2020automatic} to robustly train our models. The training hyper-parameters and training schedules were the same as the ones used in~\citet{xu2020automatic}. For evaluation, we used the Interval Bounded Propagation (IBP) method to measure verified error. Verified error represents the rate of samples within the test set that are not guaranteed to be correct within the defined robustness region regardless of the adversarial threat model. Although we use CROWN-IBP to train our models during evaluation, given its nature, adaptive verifiable training can be used with any verifiable training method. In this paper, we did not evaluate on larger-scale datasets like ImageNet~\cite{krizhevsky2012imagenet} due to the performance limitations of existing verifiable training methods on such datasets. Adaptive verifiable training can benefit from further development in verifiable training techniques that boast high performance on such datasets.

In regards to clustering the classes into groups, we first trained a non-robust model using standard cross-entropy minimization and then applied agglomerative clustering using the final layer's weights. As agglomerative clustering can be represented as a tree, we denote the root of the tree as the \textbf{top-level group split}, which splits the most dissimilar clusters apart. The extracted clusters for F-MNIST and CIFAR10 are given in the full paper.

\subsection{Comparing to CROWN-IBP}
\label{sec:results-compare}
We begin with a comparison between a baseline model trained using CROWN-IBP and our two methods: \sys (\sysshort) and Neural Decision Trees (NDT). In these experiments, our models are only trained to be robust for the given value of $\epsilon$ with respect to the top-level group split (\ie most dissimilar clusters). In F-MNIST, the top-level group split was [Trouser, Dress, Sandal, Shirt, Sneaker, Bag, Ankle Boot] vs [T-shirt/top, Pullover, Coat]. In CIFAR10, the top-level group split was [Airplane, Automobile, Ship, Truck] vs [Bird, Cat, Deer, Dog, Frog, Horse]. After the initial split, we did not impose any other robustness constraints. We denote performance with respect to the top-level split as the \textbf{inter-group error}. With respect to the NDT model, we provide results for two different architectures. The Full NDT model is a decision tree in which every node is a robust binary classifier with respect to $\epsilon$. The Mixed NDT model is a decision tree in which only the root node is robust with respect to $\epsilon$. Afterwards, all of of the inner and leaf nodes are non-robust classifiers. 

We also used two additional techniques to improve performance: upper bound scattering (UBS) and model fine tuning (FT). In upper bound scattering, instead of zeroing out the worst-case margin values of outer (or inner) group labels when calculating the inner (or outer) loss for \sysshort, we used the respective group labels' best-case margin values. Compared to \sysshort,  IGRP-UBS allows verifiable training to maintain more gradients, thus improving the estimation precision during verification. Model fine tuning, on the other hand, has been used for tasks, such as transfer learning and weight initialization. We used the baseline CROWN-IBP model or a naturally trained model (depending on the training method and robustness criteria) to initialize the weights of each node in the Neural Decision Tree. The evaluation of our models using UBS and FT is given in Table \ref{tab:improved_data}. 

\input{tables/improved_data}

Overall, we see that our approach greatly reduces the error rate of a verifiable model, while maintaining or improving the verifiable error with respect to the inter-group robustness. The performance gains increase compared to the baseline model as we: 1) increase the value of $\epsilon$ and 2) increase the complexity of the input data. With respect to the NDT models, the Full NDT-FT model has similar performance to the CROWN-IBP model. This is not surprising given that both models perform the same task, enforcing a universal robustness criteria between all classes. Once we replace all of the inner nodes in the tree with a non-robust classifier, thus prioritizing the robustness of the top-level split, the error rate of the model improves. This is especially noticeable on CIFAR10 with $\epsilon=\frac{8}{255}$, where we observe a 22.73\% error rate reduction. Finally, in our experiments, UBS and FT improves the clean performance of our models by around $1-2\%$ which means UBS and FT are effective, however, they are not necessary to improve the clean performance of the verifiably robust models. The performance of these models without UBS and FT can be found in the full paper.

\subsection{Enforcing Multiple Robustness Criteria}
We showed that our approach improves the clean performance of verifiable training by enforcing large robustness criteria on the most dissimilar classes. Here we demonstrate that our methods can further enforce multiple robustness criteria for different group splits. For that purpose, we add a new inner or outer loss term to the \sysshort loss function or change the robustness of one or more tree nodes. In Table \ref{tab:mixed_eps_data}, we present results for models trained with multiple robustness criteria. For the top-level group split, we use a large value of $\epsilon$. For the splits afterwards, in which a group contains more than two labels (\eg [Airplane, Ship, Automobile, Truck]), we use a small value of $\epsilon$. Finally, when a group split consists of only two labels (\eg [Automobile, Truck]), we do not enforce any robustness criteria.

\input{tables/mixed_eps_data}

As before, inter-group error measures the error with respect to the top-level split. We denote the error rate of groups trained using the small value of $\epsilon$ as the \textbf{intra-group error}. We note that although we employ multiple robustness criteria, our models outperform the clean performance of the baseline model trained with large $\epsilon$. Compared to the baseline CROWN-IBP model trained with large $\epsilon$, our models had a lower natural and intra-group error rate, while still maintaining inter-group and intra-group performance. Compared to the baseline CROWN-IBP model trained with small $\epsilon$, our models often had a slightly higher natural error rate, due to the need to optimize inter-group error against large $\epsilon$. Interestingly on CIFAR10, our Mixed NDT models outperformed both baseline models with respect to natural error. 

\input{tables/trunc_data}
\subsection{Truncated NDT}
In our earlier experiments, every node in the NDT was a binary classifier, which means that for a balanced tree, the depth of the tree is $\log_{2}(k)$, where k is the number of class labels. On CIFAR100, we discovered that both the Full and Mixed NDT models had a very high error rate compared to the baseline CROWN-IBP model due to the increased tree depth. The deeper a tree is, the more errors near the root of the tree will affect overall performance due to error propagation. For CIFAR10, which only had 9 binary classifiers and a tree depth of 3, these errors did not drastically hurt the overall performance. However, for CIFAR100, there were 99 binary classifier resulting in a tree depth of 7.

In addition to error propagation, high-error nodes near the bottom of the tree, likely due to a lack of data and data imbalance, contributed to the NDT's poor performance. To address both issues, we reduced the depth of the tree by merging the lower robust binary classifiers into a single non-robust classifier. If a sub-tree in the NDT contains classifiers that are all trained for the same robustness criteria, we compressed the sub-tree into a single classifier.

The final classifier has an output equal to the number of outgoing edges at the end of the original sub-tree. Preliminary experiments showed that we would achieve similar or better performance with this approach. Thus, our truncated mixed NDT is composed of two types of nodes. At the root and inner nodes, we use robust binary classifiers trained with $\epsilon=\frac{2}{255}$. At the leaf nodes, we use non-robust classifiers, created from compressing the rest of the tree. The leaf nodes determine the final classification output. Note that although the NDT is robust at more than just the root node, we still measure the inter-group error based on the top-level group split at the root. Table \ref{tab:trunc_data} presents the truncated tree results on CIFAR100.

As we see in the table, both the baseline and the Full NDT models have extremely poor performance on CIFAR100 when trained using $\epsilon=\frac{2}{255}$. We observe that by cutting the depth of the tree in half, there is a 7.75\% reduction in the error rate. Further reductions in depth improve clean performance while preserving the inter-group error, as the root node is unaffected during truncation. From these experiments, we see that there is a trade-off; reducing the depth of the tree indeed improves performance, but limits the granularity of the robustness criteria we can enforce.

%% file: tables/improved_data.tex
\begin{table*}[tbh!]
\begin{center}
\resizebox{0.85\textwidth}{!}{%
  \begin{tabular}{@{}cccccc@{}}
    \toprule
    \textbf{Dataset} & \textbf{Inter-Group $\epsilon$} & \textbf{Method} & \textbf{Error} & \textbf{Inter-Group Error} & \textbf{Verified Inter-Group Error} \\
    \midrule
    \multirow{8}{*}{F-MNIST} & \multirow{4}{*}{0.1} & CROWN-IBP & 15.40\% & 9.23\% & 14.53\% \\ 
    \cmidrule(lr){3-6}
    &  & \sysshort-UBS  & \textbf{12.86\%} & {7.94}\% & 15.59\% \\
    &  & Full NDT-FT  & 15.42\%  & 9.37\%  & 13.70\% \\
    &  & Mixed NDT-FT & \textbf{12.37\%}  & 9.37\% & 13.70\% \\ 
    \cmidrule(r){2-6}
    & \multirow{4}{*}{0.3} & CROWN-IBP & 26.22\% & 13.66\% & 23.39\% \\ 
    \cmidrule(lr){3-6}
    &  & \sysshort-UBS  & \textbf{18.82\%} & {11.02\%}  & {26.54\%}  \\ 
    &  & Full NDT-FT  & 27.04\% & 13.65\% & 21.59\%   \\
    &  & Mixed NDT-FT & \textbf{16.59\%} & {13.65\%} & {21.59\%}  \\ 
    \midrule
    \multirow{8}{*}{CIFAR10} & \multirow{4}{*}{$\cfrac{2}{255}$} & CROWN-IBP & 44.25\% & 8.62\% & 14.74\% \\ 
    \cmidrule(lr){3-6}
    &  & \sysshort-UBS  & \textbf{34.54\%} & {6.28\%} & {15.43\%} \\
    &  & Full NDT-FT  & 40.00\% & 7.25\%   & 12.31\%  \\
    &  & Mixed NDT-FT  & \textbf{22.36\%}  & {7.25\%}  & {12.31\%}  \\ 
    \cmidrule(r){2-6}
    & \multirow{4}{*}{$\cfrac{8}{255}$} & CROWN-IBP & 57.10\% & 13.77\% & 24.60\% \\
    \cmidrule(lr){3-6}
    &  & \sysshort-UBS  & \textbf{43.66\%} & {9.47\%} & {25.32\%}  \\ 
    &  & Full NDT-FT  & 58.75\%  & 11.66\% & 20.28\% \\
    &  & Mixed NDT-FT   & \textbf{26.21\%} & {11.66\%} & {20.28\%}  \\ 
    \bottomrule
  \end{tabular}
}
\caption{Performance of our \sysshort and NDT models created with Upper Bound Scattering (\sysshort-UBS) and  Fine Tuning (NDT-FT) applied. For \sysshort and NDT, the models are trained to prioritize the inter-group robustness with respect to $\epsilon$. The inter-group error is the error rate with respect to the top-level split. The verified inter-group error is the worst-case error rate within the $\epsilon$-bounded robustness region. The lower verified error is, the more robust the model is with respect to the group split.}
\label{tab:improved_data}
\end{center}
\end{table*}

%% file: tables/mixed_eps_data.tex
\begin{table*}[tbh!]
\begin{center}
\resizebox{0.9\textwidth}{!}{%
  \begin{tabular}{@{}ccccccc@{}}
    \toprule
    \textbf{Dataset} & \textbf{Inter-Group $\epsilon$} & \textbf{Intra-Group $\epsilon$} & \textbf{Method} &\textbf{Error} & \textbf{Verified Inter-Group Error} & \textbf{Verified Intra-Group Error} \\
    \midrule
    \multirow{6}{*}{F-MNIST} & \multirow{6}{*}{0.3} & \multirow{6}{*}{0.1} & CROWN-IBP (0.1) & 15.40\% & 99.98\% & 23.31\%  \\ 
    & & & CROWN-IBP (0.3) & 26.22\% & 23.39\% & 30.70\%  \\ 
    \cmidrule(r){4-7}
    & & & \sysshort & \textbf{24.11\%} & 25.61\% & 29.62\%  \\ 
    & & & \sysshort-UBS & \textbf{20.12\%} & 25.10\% & 28.52\% \\ 
    & & & Mixed NDT & \textbf{18.97\%} & 23.99\% & 26.49\% \\
    & & & Mixed NDT-FT & \textbf{19.49\%} & 21.59\% & 25.95\% \\
    \midrule
    \multirow{6}{*}{CIFAR10} & \multirow{6}{*}{$\cfrac{8}{255}$} & \multirow{6}{*}{$\cfrac{2}{255}$} & CROWN-IBP $\left(\frac{2}{255}\right)$ & 44.25\% & 58.92\% & 43.42\%  \\[2pt] 
    & & & CROWN-IBP $\left(\frac{8}{255}\right)$ & 57.10\% & 24.60\% & 46.60\%  \\ 
    \cmidrule(r){4-7}
    & & & \sysshort & \textbf{51.58\%} & 24.89\% & 45.27\% \\ 
    & & & \sysshort-UBS & \textbf{48.13\%} & 25.43\% & 44.75\% \\ 
    & & & Mixed NDT  & \textbf{38.48\%} & 21.43\% & 41.42\% \\ 
    & & & Mixed NDT-FT  & \textbf{36.79\%} & 20.28\% & 40.26\% \\ 
    \bottomrule
  \end{tabular}
}
\caption{Results when training models with multiple robustness criteria. We set the inter-group $\epsilon$ to be large as based on our hypothesis, very dissimilar groups (\eg Animals vs Vehicles in CIFAR10) should be more easily separable in the input space. Within each group composed of similar classes, we set the intra-group robustness to be small as these are the groups that are normally hard to separate. We see that both of our approaches lower the error of the model compared to the CROWN-IBP model trained trained on large epsilon only, while also having similar or better improving the inter-group robustness.}
\label{tab:mixed_eps_data}
\end{center}
\end{table*}

%% file: tables/trunc_data.tex
\begin{table*}[tbh!]
\begin{center}
\resizebox{0.85\textwidth}{!}{%
  \begin{tabular}{@{}ccccccc@{}}
    \toprule
    \textbf{Dataset} & \textbf{$\epsilon$} & \textbf{Method} & \textbf{Depth} & \textbf{Error} & \textbf{Inter-Group Error} & \textbf{Verified Inter-Group Error} \\
    \midrule
    \multirow{5}{*}{CIFAR100} & \multirow{5}{*}{$\cfrac{2}{255}$} & CROWN-IBP &  N/A & 68.98\% & 21.61\% & 38.72\% \\
   & &  Full NDT & 7 & 86.67\% & 23.71\% & 28.70\% \\
   \cmidrule(lr){3-7}
   & &  Truncated Mixed NDT & 3 & \textbf{59.23\%} & 23.71\% & 28.70\% \\
   & &  Truncated Mixed NDT & 2 & \textbf{53.94\%} & 23.71\% & 28.70\% \\
   & &  Truncated Mixed NDT & 1 & \textbf{42.66\%} & 23.71\% & 28.70\% \\
   \bottomrule
   \end{tabular}
}
\caption{Results on CIFAR100 showing the effect of truncating the NDT. The inter-group error is measured using the group split at the root node of the NDT. By reducing the depth of tree and increasing the number of outputs at the leaf node, we can maintain the inter-group error (\ie the robustness of the root node), while reducing the error rate of the overall model.}
\label{tab:trunc_data}
\end{center}
\end{table*}

%% file: section/conclusion.tex
\section{Conclusion}
\label{sec:conclusion}
Adversarial examples are a concerning vulnerability for machine learning models as these models are used in many different security and safety critical domains. As such, verifiable training provides important and necessary guarantees as to the performance of these model in adversarial scenarios. Unfortunately, the state-of-the-art verifiable training techniques still fall short as most models have poor performance on medium to large scale datasets, which makes them challenging to use in practice. Much of this performance loss can be attributed to the failure to resolve conflicts between overlapping estimated robustness regions between similar classes during training. However, is there a need to enforce the same robustness constraint on every input?

In practice, the cost of a misclassification is usually lower if two classes share a high degree of similarity. Mislabeling inputs from one class as different, but similar class may not be a very costly mistake depending on the task domain. For example, mistaking a Speed Limit sign as a Stop sign may cause the vehicle to come to a dangerous halt. However, mistaking a Speed Limit sign as a different Speed Limit sign, while a mistake, does not drastically alter the behavior of an autonomous vehicle. The vehicle still moves, albeit, at a faster or slower speed. In fact, such a mistake may occur naturally given the close visual similarity of the two classes. In such scenarios, it is more important that the model be more robust to noise that result in high cost mistakes.

In this paper, we proposed adaptive verifiable training, a new approach to verifiable training that enables current and future verifiable training techniques to train models that enforce multiple robustness criteria. Absent pre-defined class groupings, we propose using agglomerative clustering on the final layer weights of a pre-trained model to automatically subdivide the classes into groups and sub-groups of similar classes. Given two or more groups, a robustness criteria $\epsilon$ is enforced during training based on the similarity of the groups. As the similarity between groups decreases, we can enforce stricter robustness criteria. We designed two different methods to apply adaptive verifiable training. Our \sys method followed traditional verifiable training techniques and used a customized loss function to enforce multiple robustness criteria on a single model. Our Neural Decision Tree method trained multiple robust and non-robust sub-classifiers and organized them into a decision tree ensemble. We showed that both methods resulted in robust models that, compared to state-of-the-art training techniques, improved performance on non-noisy data and achieved similar verifiable performance on adversarial data, despite enforcing multiple similarity-sensitive robustness criteria.

%% file: section/ethical_impact.tex
\label{sec:ethical}
\section{Ethical Impact}

We expect positive impacts on critical applications of machine learning from the methodology proposed in this paper. It secures a machine learning model from adversarial examples, and fortifies it to make robust decisions (\ie the decisions are consistent among similar inputs), while maintaining its prediction accuracy.
We do not see a negative ethical impact as our approach does not generate adversarial examples, nor reveal weaknesses of the protected models.

%% file: section/appendix.tex

\section{Model parameters}

Here we list the number of parameters and multiply-and-accumulates (MACs) of three different model architectures, Basic, DM-Large, and ResNet18. Basic is a 4-layer LeNet model and was used to generate the results in the main text. DM-Large is a 5-layer LeNet with much wider convolutions kernels. Both of them are used in prior work for evaluating state-of-the-art verifiable training methods~\cite{zhang2019towards,gowal2018effectiveness}.
\begin{table}[!h]
    \centering
    \begin{tabular}{|c|c|r|r|} \hline
        Dataset & Models & \# of Params &  \multicolumn{1}{|c|}{MACs} \\ \hline
        \multirow{2}{*}{F-MNIST} & Basic & 275,714 & 13,257,290 \\
        & DM-Large & 864,208 & 114,632,192 \\ \hline
        \multirow{2}{*}{CIFAR10} & Basic & 337,298 & 1,385,984 \\
        & DM-Large & 17,190,602 & 150,901,760 \\ \hline
        \multirow{2}{*}{CIFAR100} & DM-Large & 17,236,772 & 150,947,840 \\
        & ResNet18 & 11,220,132 & 556,697,600\\ \hline
    \end{tabular}
    \caption{The number of parameters and multiply-and-accumulates (MACs) for different model architectures.}
    \label{tab:my_label}
\end{table}

\section{Label Splits}

\begin{figure*}[h]
     \begin{subfigure}[b]{0.5\textwidth}
         \centering
         \includegraphics[width=\textwidth]{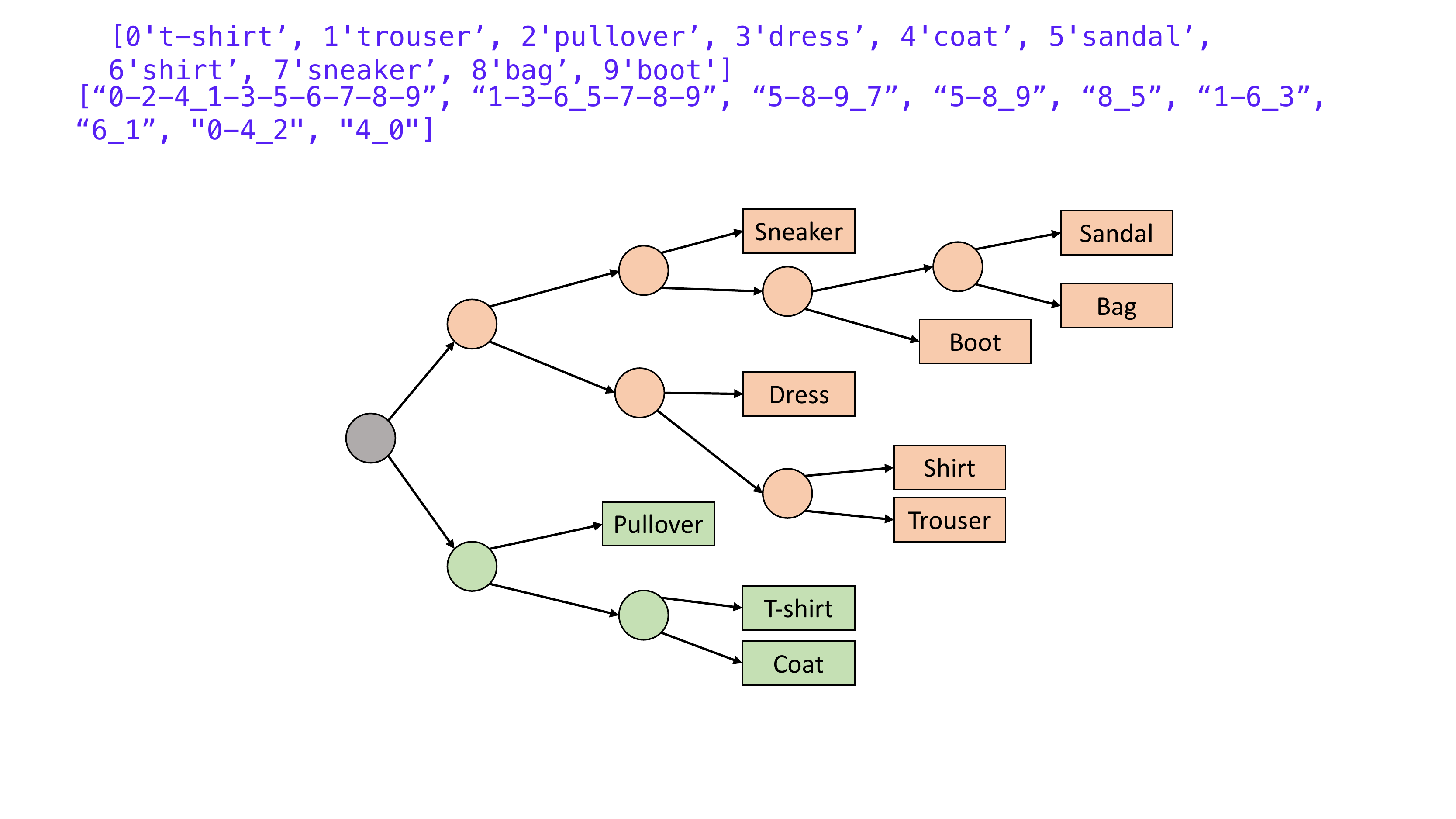}
         \caption{F-MNIST, Basic}
     \end{subfigure}
     \hfill
     \begin{subfigure}[b]{0.5\textwidth}
         \centering
         \includegraphics[width=\textwidth]{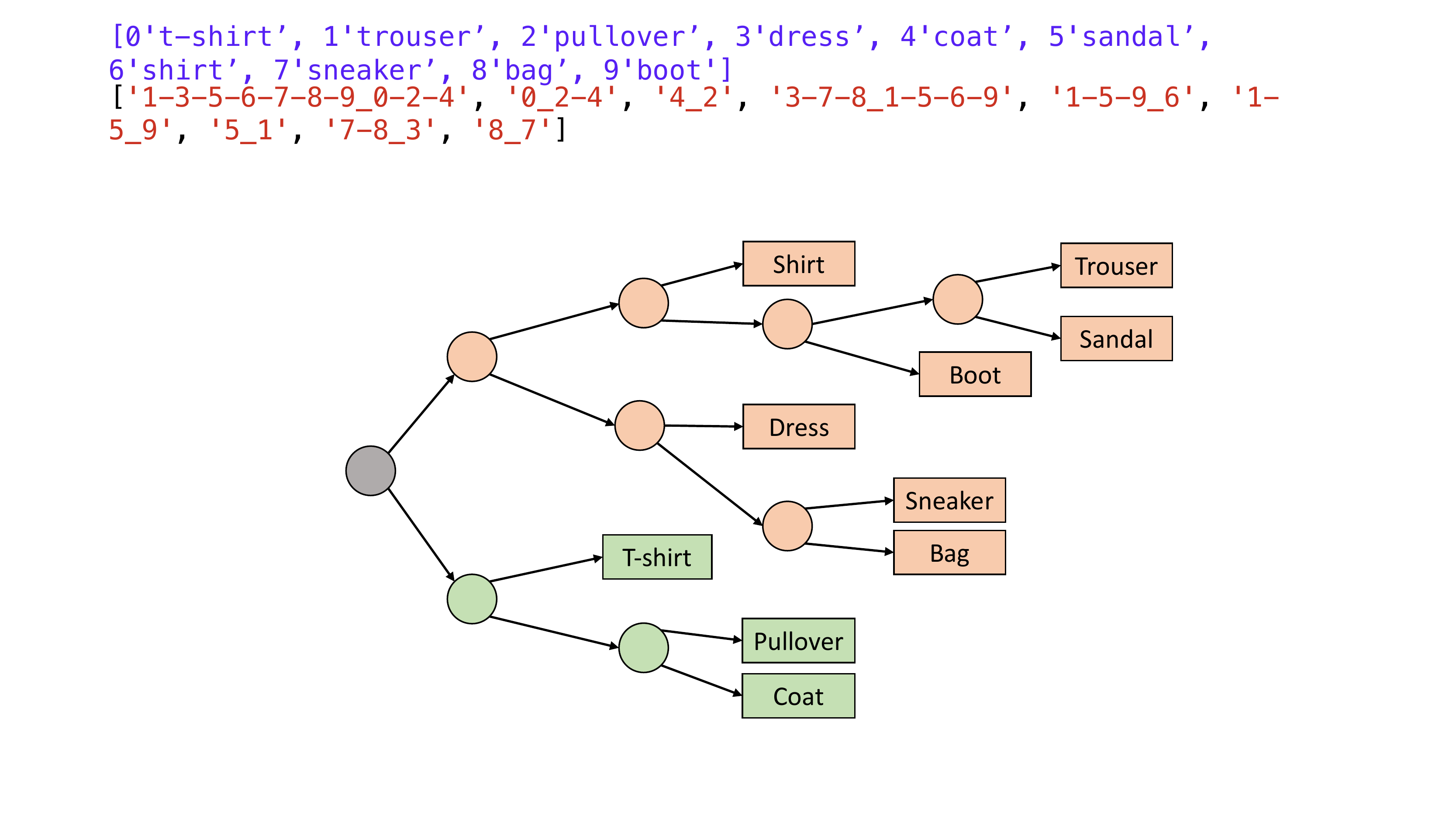}
         \caption{F-MNIST, DM-Large}
     \end{subfigure}
    \begin{subfigure}[b]{0.5\textwidth}
         \centering
         \includegraphics[width=\textwidth]{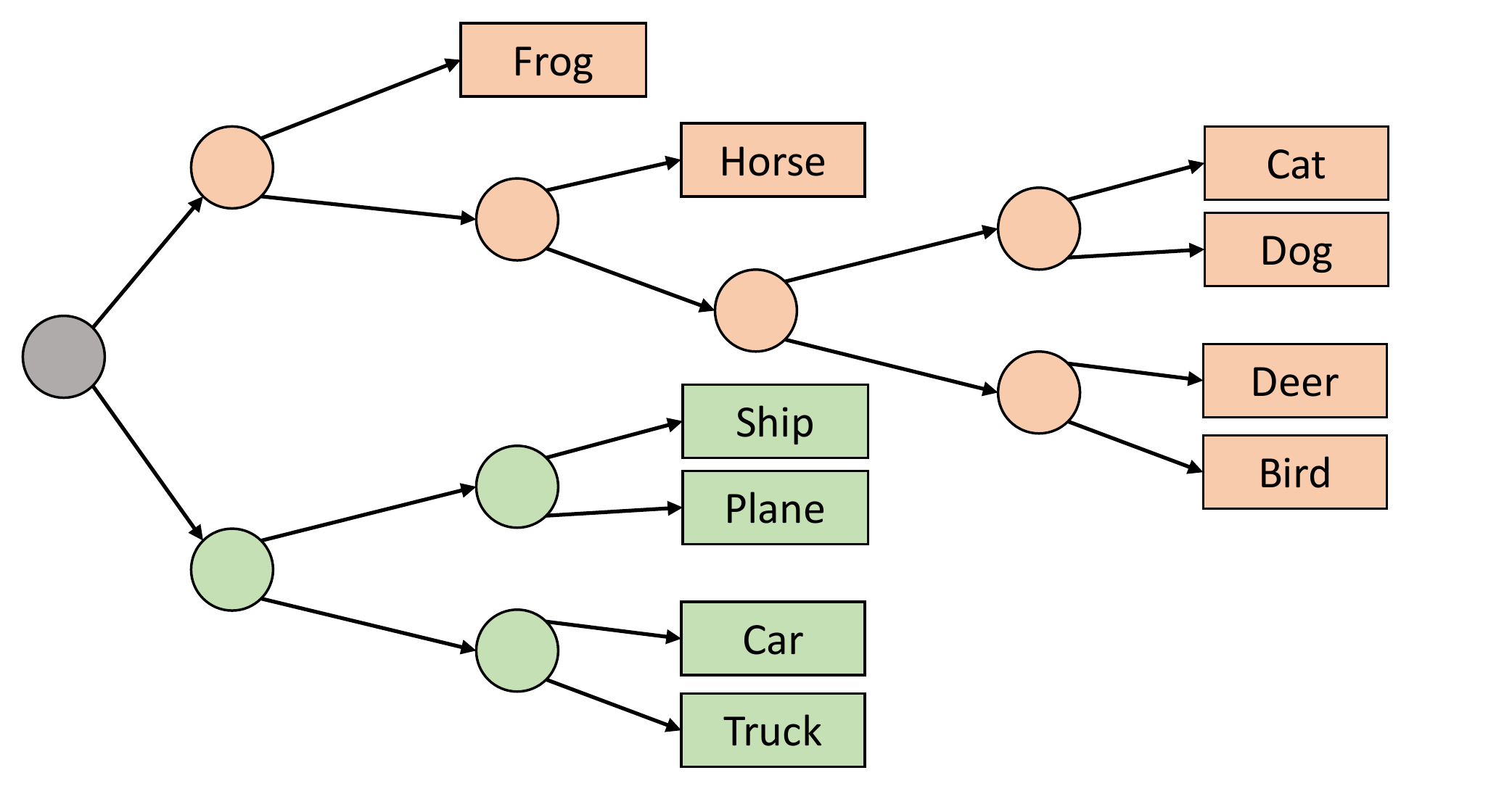}
         \caption{CIFAR10, Basic}
     \end{subfigure}
     \hfill
     \begin{subfigure}[b]{0.5\textwidth}
         \centering
         \includegraphics[width=\textwidth]{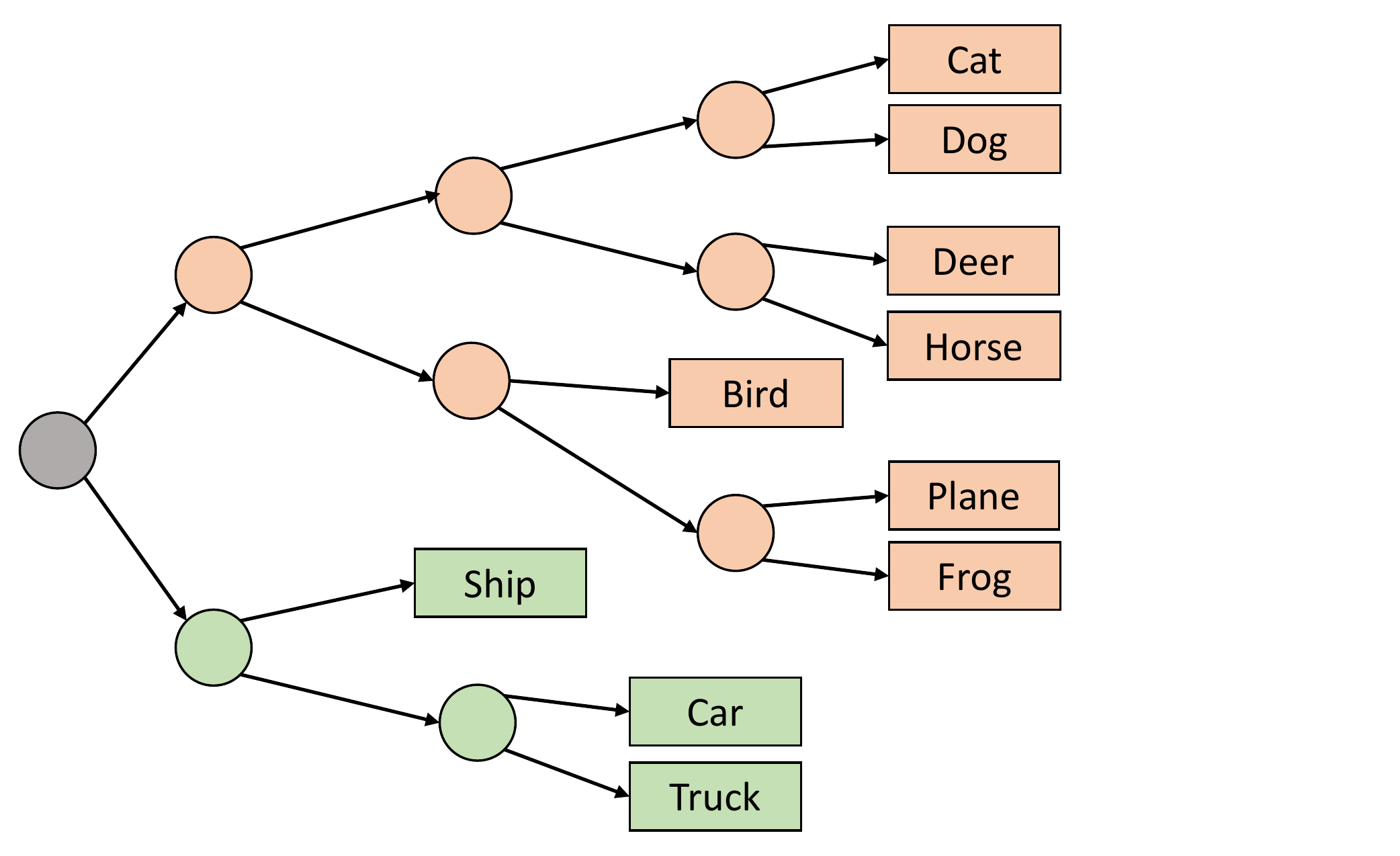}
         \caption{CIFAR10, DM-Large}
     \end{subfigure}
     \hfill

\caption{The F-MNIST and CIFAR10 label groups obtained by our class similarity identification algorithm.}
\label{fig:split_figure}
\end{figure*}

In Figure~\ref{fig:split_figure}, we provide the label splits obtained by our class similarity identification algorithms which are then used in both of our \sysshort and NDT methods.

\section{Performance of models without UBS and FT}
In Section 4.2 in the main text, we presented results of using adaptive verifiable training with upper bound scattering (UBS) and model fine tuning (FT). In Table \ref{tab:fused_data}, we present the performance of models using the same architecture and adaptive verifiable training, but without using UBS and FT. Compared to Table 1 in the main text, we see UBS and FT are not necessary to lower the error rate of our models compared to CROWN-IBP.

\input{tables/fused_data}

\section{Comparing against CROWN-IBP using DM-Large}
The basic performance of the DM-Large created using \sysshort and NDT is given in Table~\ref{tab:sup_fused_data}. The performance of the DM-Large models created using \sysshort-UBS and NDT-FT is given in Table~\ref{tab:sup_improved_data}. Compared to the CROWN-IBP, our approach still greatly improves the clean performance of the model, while maintaining or improving the inter-group error. We also see that our results are much better on CIFAR10, compared to F-MNIST, which we attribute to the increased complexity of the dataset. The increased complexity increases the hardness of separating similar classes during verification, so our approach, which accounts for similarity, naturally improves in performance compared to the baseline model. When UBS and FT are applied, we see a slight improvment in performance, especially on CIAFR10. Finally, we note that compared to results in Tables 1 in the main text and Table~\ref{tab:fused_data} here, increasing the capacity of the model does not appear to greatly impact the performance difference between the baseline model and adaptive verifiable training. 


\input{tables/sup_fused_data}

\input{tables/sup_improved_data}

%% file: tables/fused_data.tex
\begin{table*}[tbh!]
\begin{center}
\resizebox{0.85\textwidth}{!}{%
  \begin{tabular}{@{}cccccc@{}}
    \toprule
    \textbf{Dataset} & \textbf{Inter-Group $\epsilon$} & \textbf{Method} & \textbf{Error} & \textbf{Inter-Group Error} & \textbf{Verified Inter-Group Error} \\
    \midrule
    \multirow{8}{*}{F-MNIST} & \multirow{4}{*}{0.1} & CROWN-IBP & 15.40\% & 9.23\% & 14.53\% \\ 
    \cmidrule(lr){3-6}
    &  & \sysshort  & \textbf{15.21\%} & 9.81\% & 14.83\% \\
    &  & Full NDT  & 16.47\% & 9.74\% & 14.05\% \\
    &  & Mixed NDT & \textbf{12.74\%} & 9.74\% & {14.05}\% \\ 
    \cmidrule(r){2-6}
    & \multirow{4}{*}{0.3} & CROWN-IBP & 26.22\% & 13.66\% & 23.39\% \\ 
    \cmidrule(lr){3-6}
    &  & \sysshort  & \textbf{22.55\%} & {13.40\%} & {25.43\%} \\ 
    &  & Full NDT  & \textbf{25.36\%} & 12.44\% & 23.99\% \\
    &  & Mixed NDT & \textbf{15.29\%} & {12.44\%} & {23.99\%} \\ 
    \midrule
    \multirow{8}{*}{CIFAR10} & \multirow{4}{*}{$\cfrac{2}{255}$} & CROWN-IBP & 44.25\% & 8.62\% & 14.74\% \\ 
    \cmidrule(lr){3-6}
    &  & \sysshort  & \textbf{36.36\%} & {7.18\%} & {14.59\%}\\
    &  & Full NDT  & \textbf{41.06\%} & 8.02\% & 12.48\% \\
    &  & Mixed NDT  & \textbf{24.94\%} & {8.02\%} & {12.48\%} \\ 
    \cmidrule(r){2-6}
    & \multirow{4}{*}{$\cfrac{8}{255}$} & CROWN-IBP & 57.10\% & 13.77\% & 24.60\% \\
    \cmidrule(lr){3-6}
    &  & \sysshort  & \textbf{48.56\%} & {12.13\%} & {24.41\%} \\ 
    &  & Full NDT  & 61.10\% & 12.48\% & 21.43\% \\
    &  & Mixed NDT  & \textbf{28.97\%} & {12.48\%} & {21.43\%} \\ 
    \bottomrule
  \end{tabular}
}
\caption{ Performance of our \sysshort and NDT models without using upper bound scattering and fine tuning on the basic model architecture. Overall, upper bound scattering and fine tuning are not necessary to lower the error rate of our models compared to CROWN-IBP.}
\label{tab:fused_data}
\end{center}
\end{table*}

%% file: tables/sup_fused_data.tex
\begin{table*}[tbh!]
\begin{center}
\resizebox{0.85\textwidth}{!}{%
  \begin{tabular}{@{}cccccc@{}}
    \toprule
    \textbf{Dataset} & \textbf{$\epsilon$} & \textbf{Training Method} & \textbf{Error} & \textbf{Inter-Group Error} & \textbf{Verified Inter-Group Error} \\
    \midrule
    \multirow{8}{*}{F-MNIST} & \multirow{4}{*}{0.1} & CROWN-IBP & 15.10\% & {9.07}\% & 17.80\% \\ 
    \cmidrule(lr){3-6}
    &  & \sysshort  & \textbf{14.92\%} & 9.88\% & 17.86\% \\
    &  & Full NDT  & 16.01\% & 10.01\% & 14.98\% \\
    &  & Mixed NDT & \textbf{12.35\%} & 10.01\% & {14.98}\% \\ 
    \cmidrule(r){2-6}
    & \multirow{4}{*}{0.3} & CROWN-IBP & 26.12\% & {13.27}\% & 23.99\% \\ 
    \cmidrule(lr){3-6}
    &  & \sysshort  & \textbf{24.33\%} & 14.09\% & 24.82\% \\ 
    &  & Full NDT  & 29.30\% & 16.14\% & 23.56\% \\
    &  & Mixed NDT & \textbf{18.53\%} & 16.14\% & {23.56\%} \\ 
    \midrule
    \multirow{8}{*}{CIFAR10} & \multirow{4}{*}{2/255} & CROWN-IBP & 38.03\% & 7.22\% & 15.15\% \\ 
    \cmidrule(lr){3-6}
    &  & \sysshort  & \textbf{30.97}\% & {5.88}\% & 14.51\%\\
    &  & Full NDT  & 39.12\% & 7.42\% & 12.33\% \\
    &  & Mixed NDT  & \textbf{18.41\%} & 7.42\% & {12.33\%} \\ 
    \cmidrule(r){2-6}
    & \multirow{4}{*}{8/255} & CROWN-IBP & 57.48\% & 14.82\% & 27.12\% \\
    \cmidrule(lr){3-6}
    &  & \sysshort  & \textbf{46.02}\% & {13.05}\% & 29.79\% \\ 
    &  & Full NDT  & 59.26\% & 12.54\% & 21.41\% \\
    &  & Mixed NDT  & \textbf{23.02\%} & {12.54\%} & {21.41\%} \\ 
    \bottomrule
  \end{tabular}
}
\caption{Performance of our \sysshort and NDT models using the DM-Large architecture. Both \sysshort and NDT improve clean performance compared to the baseline model.}
\label{tab:sup_fused_data}
\end{center}
\end{table*}

%% file: tables/sup_improved_data.tex
\begin{table*}[tbh!]
\begin{center}
\resizebox{0.85\textwidth}{!}{%
  \begin{tabular}{@{}cccccc@{}}
    \toprule
    \textbf{Dataset} & \textbf{Inter-Group $\epsilon$} & \textbf{Method} & \textbf{Error} & \textbf{Inter-Group Error} & \textbf{Verified Inter-Group Error} \\
    \midrule
    \multirow{8}{*}{F-MNIST} & \multirow{4}{*}{0.1} & CROWN-IBP & 15.10\% & 9.07\% & 17.80\% \\ 
    \cmidrule(lr){3-6}
    &  & \sysshort-UBS  & \textbf{12.97\%}  & {8.12}\%  & 19.09\% \\
    &  & Full NDT-FT  & 16.03\% & 10.26\% & 15.56\% \\
    &  & Mixed NDT-FT & \textbf{12.79\%} & 10.26\% & 15.56\% \\ 
    \cmidrule(r){2-6}
    & \multirow{4}{*}{0.3} & CROWN-IBP & 26.12\% & 13.27\% & 23.99\% \\ 
    \cmidrule(lr){3-6}
    &  & \sysshort-UBS  & \textbf{21.60\%} & {12.93\%} & {24.98\%} \\ 
    &  & Full NDT-FT  & 29.47\% & 16.37\% & 23.40\% \\
    &  & Mixed NDT-FT & \textbf{19.00\%} & {16.37\%} & {23.40\%}  \\ 
    \midrule
    \multirow{8}{*}{CIFAR10} & \multirow{4}{*}{$\cfrac{2}{255}$} & CROWN-IBP & 38.03\% & 7.22\% & 15.15\% \\ 
    \cmidrule(lr){3-6}
    &  & \sysshort-UBS  & \textbf{29.96\%} &  5.88\% &  14.51\%  \\
    &  & Full NDT-FT  & 35.72\% & 6.29\%  & 11.47\%  \\
    &  & Mixed NDT-FT  & \textbf{14.32\%} & 6.29\%   & 11.47\% \\ 
    \cmidrule(r){2-6}
    & \multirow{4}{*}{$\cfrac{8}{255}$} & CROWN-IBP & 57.48\% & 14.82\% & 27.12\% \\
    \cmidrule(lr){3-6}
    &  & \sysshort-UBS  & \textbf{41.41\%}  & 10.15\% & 24.43\%  \\ 
    &  & Full NDT-FT  & 58.12\%  & 12.25\%  & 20.64\% \\
    &  & Mixed NDT-FT   & \textbf{20.04\%} & 12.25\% & 20.64\%  \\ 
    \bottomrule
  \end{tabular}
}
\caption{ Performance of our \sysshort-UBS and NDT-FT models using the DM-Large architecture. Compared to Table \ref{tab:sup_fused_data}, we see that UBS and FT slightly improve the performance of our models.}
\label{tab:sup_improved_data}
\end{center}
\end{table*}

%% file: ref.bbl
\begin{thebibliography}{34}
\providecommand{\natexlab}[1]{#1}
\providecommand{\url}[1]{\texttt{#1}}
\providecommand{\urlprefix}{URL }
\expandafter\ifx\csname urlstyle\endcsname\relax
  \providecommand{\doi}[1]{doi:\discretionary{}{}{}#1}\else
  \providecommand{\doi}{doi:\discretionary{}{}{}\begingroup
  \urlstyle{rm}\Url}\fi

\bibitem[{Athalye, Carlini, and Wagner(2018)}]{athalye2018obfuscated}
Athalye, A.; Carlini, N.; and Wagner, D. 2018.
\newblock Obfuscated gradients give a false sense of security: Circumventing
  defenses to adversarial examples.
\newblock In \emph{International Conference on Machine Learning (ICML)}.

\bibitem[{Balunovic and Vechev(2020)}]{balunovic2020Adversarial}
Balunovic, M.; and Vechev, M. 2020.
\newblock Adversarial Training and Provable Defenses: Bridging the Gap.
\newblock In \emph{International Conference on Learning Representations
  (ICLR)}.

\bibitem[{Bhagoji et~al.(2018)Bhagoji, He, Li, and
  Song}]{Bhagoji2018PracticalBA}
Bhagoji, A.~N.; He, W.; Li, B.; and Song, D.~X. 2018.
\newblock Practical Black-Box Attacks on Deep Neural Networks Using Efficient
  Query Mechanisms.
\newblock In \emph{European Conference on Computer Vision (ECCV)}.

\bibitem[{Chen et~al.(2017)Chen, Zhang, Sharma, Yi, and Hsieh}]{zoo}
Chen, P.-Y.; Zhang, H.; Sharma, Y.; Yi, J.; and Hsieh, C.-J. 2017.
\newblock ZOO: Zeroth Order Optimization Based Black-box Attacks to Deep Neural
  Networks Without Training Substitute Models.
\newblock In \emph{ACM Workshop on Artificial Intelligence and Security
  (AiSec)}.

\bibitem[{Cheng et~al.(2018)Cheng, Le, Chen, Zhang, Yi, and
  Hsieh}]{Cheng2018QueryEfficientHB}
Cheng, M.; Le, T.; Chen, P.-Y.; Zhang, H.; Yi, J.; and Hsieh, C.-J. 2018.
\newblock Query-Efficient Hard-label Black-box Attack: An Optimization-based
  Approach.
\newblock \emph{arXiv:1807.04457} .

\bibitem[{Cohen, Rosenfeld, and Kolter(2019)}]{cohen2019certified}
Cohen, J.~M.; Rosenfeld, E.; and Kolter, J.~Z. 2019.
\newblock Certified adversarial robustness via randomized smoothing.
\newblock \emph{International Conference on Machine Learning (ICML)} .

\bibitem[{Fischer et~al.(2019)Fischer, Balunovic, Drachsler-Cohen, Gehr, Zhang,
  and Vechev}]{fischer2019dl2}
Fischer, M.; Balunovic, M.; Drachsler-Cohen, D.; Gehr, T.; Zhang, C.; and
  Vechev, M. 2019.
\newblock Dl2: Training and querying neural networks with logic.
\newblock In \emph{International Conference on Machine Learning (ICML)}.

\bibitem[{Goodfellow, Shlens, and Szegedy(2014)}]{goodfellow2014explaining}
Goodfellow, I.~J.; Shlens, J.; and Szegedy, C. 2014.
\newblock Explaining and harnessing adversarial examples.
\newblock In \emph{International Conference on Learning Representations
  (ICLR)}.

\bibitem[{Gowal et~al.(2019)Gowal, Dvijotham, Stanforth, Bunel, Qin, Uesato,
  Arandjelovic, Mann, and Kohli}]{gowal2018effectiveness}
Gowal, S.; Dvijotham, K.; Stanforth, R.; Bunel, R.; Qin, C.; Uesato, J.;
  Arandjelovic, R.; Mann, T.; and Kohli, P. 2019.
\newblock Scalable Verified Training for Provably Robust Image Classification.
\newblock In \emph{International Conference on Computer Vision (ICCV)}.

\bibitem[{Jonas and Evans(2020)}]{jonas2020certifying}
Jonas, M.~A.; and Evans, D. 2020.
\newblock Certifying Joint Adversarial Robustness for Model Ensembles.
\newblock \emph{arXiv:2004.10250} .

\bibitem[{Krizhevsky, Sutskever, and Hinton(2012)}]{krizhevsky2012imagenet}
Krizhevsky, A.; Sutskever, I.; and Hinton, G.~E. 2012.
\newblock Imagenet classification with deep convolutional neural networks.
\newblock In \emph{Advances in Neural Information Processing Systems
  (NeurIPS)}.

\bibitem[{Kurakin, Goodfellow, and Bengio(2016)}]{kurakin2016adversarial}
Kurakin, A.; Goodfellow, I.; and Bengio, S. 2016.
\newblock Adversarial examples in the physical world.
\newblock \emph{arXiv:1607.02533} .

\bibitem[{Liu et~al.(2019)Liu, Arnon, Lazarus, Barrett, and
  Kochenderfer}]{liu2019algorithms}
Liu, C.; Arnon, T.; Lazarus, C.; Barrett, C.; and Kochenderfer, M.~J. 2019.
\newblock Algorithms for verifying deep neural networks.
\newblock \emph{arXiv:1903.06758} .

\bibitem[{Liu et~al.(2016)Liu, Chen, Liu, and Song}]{liu2016delving}
Liu, Y.; Chen, X.; Liu, C.; and Song, D. 2016.
\newblock Delving into transferable adversarial examples and black-box attacks.
\newblock In \emph{International Conference on Learning Representations
  (ICLR)}.

\bibitem[{Mirman, Gehr, and Vechev(2018)}]{mirman2018differentiable}
Mirman, M.; Gehr, T.; and Vechev, M. 2018.
\newblock Differentiable abstract interpretation for provably robust neural
  networks.
\newblock In \emph{International Conference on Machine Learning (ICML)}.

\bibitem[{Papernot, McDaniel, and
  Goodfellow(2016)}]{papernot2016transferability}
Papernot, N.; McDaniel, P.; and Goodfellow, I. 2016.
\newblock Transferability in machine learning: from phenomena to black-box
  attacks using adversarial samples.
\newblock \emph{arXiv:1605.07277} .

\bibitem[{Papernot et~al.(2016)Papernot, McDaniel, Jha, Fredrikson, Celik, and
  Swami}]{papernot2016limitations}
Papernot, N.; McDaniel, P.; Jha, S.; Fredrikson, M.; Celik, Z.~B.; and Swami,
  A. 2016.
\newblock The limitations of deep learning in adversarial settings.
\newblock In \emph{IEEE European Symposium on Security and Privacy (EuroS\&P)}.

\bibitem[{Raghunathan, Steinhardt, and Liang(2018)}]{raghunathan2018certified}
Raghunathan, A.; Steinhardt, J.; and Liang, P. 2018.
\newblock Certified defenses against adversarial examples.
\newblock In \emph{International Conference on Learning Representations
  (ICLR)}.

\bibitem[{Salman et~al.(2019{\natexlab{a}})Salman, Li, Razenshteyn, Zhang,
  Zhang, Bubeck, and Yang}]{salman2019provably}
Salman, H.; Li, J.; Razenshteyn, I.; Zhang, P.; Zhang, H.; Bubeck, S.; and
  Yang, G. 2019{\natexlab{a}}.
\newblock Provably robust deep learning via adversarially trained smoothed
  classifiers.
\newblock In \emph{Advances in Neural Information Processing Systems
  (NeurIPS)}.

\bibitem[{Salman et~al.(2019{\natexlab{b}})Salman, Yang, Zhang, Hsieh, and
  Zhang}]{salman2019convex}
Salman, H.; Yang, G.; Zhang, H.; Hsieh, C.-J.; and Zhang, P.
  2019{\natexlab{b}}.
\newblock A convex relaxation barrier to tight robust verification of neural
  networks.
\newblock In \emph{Advances in Neural Information Processing Systems
  (NeurIPS)}.

\bibitem[{Singh et~al.(2018)Singh, Gehr, Mirman, P{\"u}schel, and
  Vechev}]{singh2018fast}
Singh, G.; Gehr, T.; Mirman, M.; P{\"u}schel, M.; and Vechev, M. 2018.
\newblock Fast and effective robustness certification.
\newblock In \emph{Advances in Neural Information Processing Systems
  (NeurIPS)}.

\bibitem[{Szegedy et~al.(2014)Szegedy, Zaremba, Sutskever, Bruna, Erhan,
  Goodfellow, and Fergus}]{szegedy2014intriguing}
Szegedy, C.; Zaremba, W.; Sutskever, I.; Bruna, J.; Erhan, D.; Goodfellow, I.;
  and Fergus, R. 2014.
\newblock Intriguing properties of neural networks.
\newblock In \emph{International Conference on Learning Representations
  (ICLR)}.

\bibitem[{Wan et~al.(2020)Wan, Dunlap, Ho, Yin, Lee, Jin, Petryk, Bargal, and
  Gonzalez}]{wan2020nbdt}
Wan, A.; Dunlap, L.; Ho, D.; Yin, J.; Lee, S.; Jin, H.; Petryk, S.; Bargal,
  S.~A.; and Gonzalez, J.~E. 2020.
\newblock NBDT: Neural-Backed Decision Trees.
\newblock \emph{arXiv:2004.00221} .

\bibitem[{Wang et~al.(2018{\natexlab{a}})Wang, Chen, Abdou, and
  Jana}]{wang2018mixtrain}
Wang, S.; Chen, Y.; Abdou, A.; and Jana, S. 2018{\natexlab{a}}.
\newblock Mixtrain: Scalable training of formally robust neural networks.
\newblock \emph{arXiv:1811.02625} .

\bibitem[{Wang et~al.(2018{\natexlab{b}})Wang, Pei, Whitehouse, Yang, and
  Jana}]{wang2018efficient}
Wang, S.; Pei, K.; Whitehouse, J.; Yang, J.; and Jana, S. 2018{\natexlab{b}}.
\newblock Efficient formal safety analysis of neural networks.
\newblock In \emph{Advances in Neural Information Processing Systems
  (NeurIPS)}.

\bibitem[{Weng et~al.(2018)Weng, Zhang, Chen, Song, Hsieh, Boning, Dhillon, and
  Daniel}]{weng2018towards}
Weng, T.-W.; Zhang, H.; Chen, H.; Song, Z.; Hsieh, C.-J.; Boning, D.; Dhillon,
  I.~S.; and Daniel, L. 2018.
\newblock Towards fast computation of certified robustness for relu networks.
\newblock In \emph{International Conference on Machine Learning (ICML)}.

\bibitem[{Wong and Kolter(2018)}]{wong2018provable}
Wong, E.; and Kolter, Z. 2018.
\newblock Provable defenses against adversarial examples via the convex outer
  adversarial polytope.
\newblock In \emph{International Conference on Machine Learning (ICML)}.

\bibitem[{Wong et~al.(2018)Wong, Schmidt, Metzen, and Kolter}]{wong2018scaling}
Wong, E.; Schmidt, F.; Metzen, J.~H.; and Kolter, J.~Z. 2018.
\newblock Scaling provable adversarial defenses.
\newblock In \emph{Advances in Neural Information Processing Systems
  (NeurIPS)}.

\bibitem[{Xu et~al.(2020{\natexlab{a}})Xu, Shi, Zhang, Huang, Chang, Kailkhura,
  Lin, and Hsieh}]{xu2020automatic}
Xu, K.; Shi, Z.; Zhang, H.; Huang, M.; Chang, K.-W.; Kailkhura, B.; Lin, X.;
  and Hsieh, C.-J. 2020{\natexlab{a}}.
\newblock Automatic Perturbation Analysis on General Computational Graphs.
\newblock \emph{arXiv:2002.12920} .

\bibitem[{Xu et~al.(2020{\natexlab{b}})Xu, Shi, Zhang, Wang, Chang, Huang,
  Kailkhura, Lin, and Hsieh}]{xu2020provable}
Xu, K.; Shi, Z.; Zhang, H.; Wang, Y.; Chang, K.-W.; Huang, M.; Kailkhura, B.;
  Lin, X.; and Hsieh, C.-J. 2020{\natexlab{b}}.
\newblock Provable, Scalable and Automatic Perturbation Analysis on General
  Computational Graphs.
\newblock \emph{arXiv:2002.12920} .

\bibitem[{Zhang et~al.(2020)Zhang, Chen, Xiao, Gowal, Stanforth, Li, Boning,
  and Hsieh}]{zhang2019towards}
Zhang, H.; Chen, H.; Xiao, C.; Gowal, S.; Stanforth, R.; Li, B.; Boning, D.;
  and Hsieh, C.-J. 2020.
\newblock Towards stable and efficient training of verifiably robust neural
  networks.
\newblock In \emph{International Conference on Learning Representations
  (ICLR)}.

\bibitem[{Zhang, Cheng, and Hsieh(2019)}]{zhang2019enhancing}
Zhang, H.; Cheng, M.; and Hsieh, C.-J. 2019.
\newblock Enhancing Certifiable Robustness via a Deep Model Ensemble.
\newblock \emph{arXiv:1910.14655} .

\bibitem[{Zhang et~al.(2018)Zhang, Weng, Chen, Hsieh, and
  Daniel}]{zhang2018efficient}
Zhang, H.; Weng, T.-W.; Chen, P.-Y.; Hsieh, C.-J.; and Daniel, L. 2018.
\newblock Efficient neural network robustness certification with general
  activation functions.
\newblock In \emph{Advances in neural information processing systems
  (NeurIPS)}.

\bibitem[{Zhang and Evans(2019)}]{zhang2018cost}
Zhang, X.; and Evans, D. 2019.
\newblock Cost-sensitive robustness against adversarial examples.
\newblock In \emph{International Conference on Learning Representations
  (ICLR)}.

\end{thebibliography}
